# Predicting housing prices and analyzing real estate markets in the Chicago suburbs using machine learning

Kevin Xu[1], Hieu Nguyen[2]
[1]Neuqua Valley High School, Naperville, Illinois
[2]Mentor, University of Connecticut, Storrs, CT

ABSTRACT

The pricing of housing properties is determined by a variety of factors. However, post-pandemic markets have experienced volatility in the Chicago suburb area, which have affected house prices greatly. In this study, analysis was done on the Naperville/Bolingbrook real estate market to predict property prices based on these housing attributes through machine learning models, and to evaluate the effectiveness of such models in a volatile market space. Gathering data from Redfin, a real estate website, sales data from 2018 up until the summer season of 2022 were collected for research. By analyzing these sales in this range of time, we can also look at the state of the housing market and identify trends in price. For modeling the data, the models used were linear regression, support vector regression, decision tree regression, random forest regression, and XGBoost regression. To analyze results, comparison was made on the MAE, RMSE, and R-squared values for each model. It was found that the XGBoost model performs the best in predicting house prices despite the additional volatility sponsored by post-pandemic conditions. After modeling, Shapley Values (SHAP) were used to evaluate the weights of the variables in constructing models. The code and data files can be found at https://github.com/ GeometricBison/HousePriceML.

## Introduction

In real estate markets, appraisals of home value are essential for conducting business. To estimate the price of a property, many details about the property must be considered. Attributes like square footage or number of rooms can easily sway the price of a house up or down. The goal of this project is to construct a machine learning model that can estimate the price of a property in suburban Chicago through evaluating its attributes. For example, in most cases, an increase in square footage would certainly increase the price of a property. However, diving deeper, not all variables are created equal. The model would have to consider the different weights of the variables and consider whether one variable would be more influential to the price than another. Another goal of this project is to analyze the real estate market by using models to identify trends. To accomplish this, models were constructed for different years from 2018 up until July of 2022. For example, in a volatile housing market year like 2022, prices have shot up tremendously as demand for homes becomes higher and higher. However, comparison to more stable markets before the pandemic can reveal insights on market volatility and whether prices are high or low. By comparing models, trends can be analyzed over the years for the real estate market.

    In order to conduct this project, three major steps were involved in the process. Data had to be extracted from Redfin.com, a realtor database website, the data then had to be transformed and analyzed, and finally a model would have to be applied to the newly formed data to estimate the price. A script was used to load pages from Redfin, parse through its contents to find relative data, and to write the contents to an excel file which held all the data in an organized manner. This was all possible due to the requests-html library, a Python package that specializes in rendering JavaScript based websites for data. Data was then categorized by its sell date and city location. Analysis was done on the data, which included how important the attribute was for the housing price and transforming categorical data points into a numerical format which could be interpreted by the model.

    Currently, machine learning has proved to itself to be very useful in this market sphere, as shown by companies employing these techniques like Zillow's Zestimate and Redfin's home appraisal. In this study, analysis was done with models like support vector regression, XGBoost models, and random forest regression. Comparisons between model effectiveness can be detailed in the results section. The organization of the paper is as follows: section 2 will go over data collection, section 3 will discuss data preprocessing, section 4 will describe data analysis, section 5 will cover modeling, section 6 will detail the results of the study, and section 7 will describe variable importance.

## Data Collection

Python was used to scrape the data off Redfin from listings in the cities of Bolingbrook and Naperville. The requests-html library was used specifically for this website. Since Redfin is a modern website and uses JavaScript to render its pages, a more powerful web scraper library had to be used to parse the data. This required to render the website first, and then extract the complete HTML code. First, an array of links was established that held all the pages of listings on the website. The URLs to these pages were formatted with a base URL and a page number, so individual listings had to be found through each page in the array. Using the web scraper, the html data was parsed to identify each link to the properties displayed on the page, which were again collected and stored. Given the large volume of links to render and scrape, the code was constructed to run asynchronously. On each page, data collection was relatively simple. All Redfin listings share a similar format, so by analyzing the patterns in the HTML and CSS of one Redfin page, it could be applied to other pages. However, sometimes homeowners do not put all data on their listings, so some data points are empty. Additional steps required to process this data will be discussed in the next section. The attributes that were scraped consist of: square footage, property type, year built, price, number of car spaces, address, high school, beds, baths (half and full), heating, cooling, number of carpet rooms, number of hardwood rooms, basement, basement square footage, basement description, and tax annual amount. Entries that did not have the data point given were simply put in as null. These data entries are stored in CSV files categorized by city and name.

## Data Preprocessing

The CSV data had to be cleaned for further analysis. The variables collected were square footage, property type, year built, price, number of car spaces, address, high school, beds, baths (half and full), heating, cooling, number of carpet rooms, number of hardwood rooms, basement, basement square footage, basement description, and tax annual amount. The variable square footage is very important for a property. Data transformation was done by only removing the dollar sign and turning it into an integer. For property type, there were three categories: townhouse, condo/co-op, and single residential. To make analysis easier, additional categories were made for categorical variables like home type. Each column had a 0 or 1 indicating whether a property was a certain type. For example, a condo would have a 1 in the condo column and a zero in both the single residential and townhouse columns. Another two important metrics were bedrooms and bathrooms. Bathrooms specifically were separated into two categories: half and full. Heating, cooling, and basement description were all treated as categorical variables, where the operation that was applied to the property type variable was also applied to these categories as well. However, these categories are not binomial categories but rather multinomial features, thus requiring additional processing.

Using regex, a library that specializes in parsing text, separation of broader categories was done. For example, heating had a variety of descriptions and combinations with common terms, like "natural gas." Therefore, three main categories were made up instead: natural gas, baseboard, and an "other" category. For cooling, the columns were zoned, central air, and "other" once again. For the basement variable, there were a lot more, but the main ones once again were none, full, partial, English, and walk-out. Number of carpet and number of hardwood rooms were also useful in this analysis. They were combined to make a number of rooms columns to avoid bias, as a zero-carpet house could still hold high value if it had more hardwood floors instead. With these transformed variables, some outliers could be removed to improve the data. For example, properties with a value of $2,500,000 or greater were removed. Properties were also limited to a ten thousand square foot limit. Finally, any homes without a listing price were removed, as it would interfere in the modeling process. The table below depicts the averages of the housing attributes as organized by year and variable.

## Data Analysis

From table 1, it can be seen that over the course of the past four years, most of the variables remain the same in averages. The one exception to this rule is the property price, proving a hot market trend for the real estate market in Naperville. The largest jump in price occurred in the past year (2021-2022). This drastic increase in price variation also resulted in slightly more volatility as described below. Using a correlation function, the most important features were analyzed. Tax annual amount and square footage appeared as two strong variables in price prediction. Single family residential homes supported a higher house price and condos supported a strong lower correlation. Bathrooms were more important than bedrooms in terms of correlation as well, especially the number of full bathrooms. Figure 1 shows a heatmap of all notable variables.

|  | Year Sold | | | |
|---|---|---|---|---|
| Attribute | 2018 | 2019 | 2020 | 2021-2022 |
| Year Built | 1988.65 | 1988.97 | 1989.86 | 1989.19 |
| Price | 335097.5 | 355088.4 | 386857.5 | 455886 |
| Car Spaces | 2.09 | 2.11 | 2.16 | 2.09 |
| Beds | 3.46 | 3.5 | 3.52 | 3.29 |
| Baths | 2.55 | 2.59 | 2.67 | 2.56 |
| Sqft | 2184.67 | 2275.22 | 2348.03 | 2292.89 |
| Full Baths | 2.19 | 2.25 | 2.34 | 2.31 |
| Half Baths | 1.05 | 1.04 | 1.05 | 1.12 |
| Carpet Rooms | 4.98 | 5.24 | 5.1 | 4.97 |
| Hardwood Rooms | 4.67 | 4.76 | 4.92 | 4.43 |
| Total Number of Rooms | 8.29 | 8.47 | 8.55 | 8.01 |
| Basement Sqft | 1259.53 | 1467.43 | 1185.96 | 1312.58 |
| Tax Annual Amount | 7494.57 | 8050.56 | 8537.79 | 8545.48 |

**Table 1.** Averages of Housing Data
Details about property attributes can be seen. Most categories stay the same throughout the years, though the price and tax amount experience a gradual increase through the years 2018-2020 and explode in value after 2021.

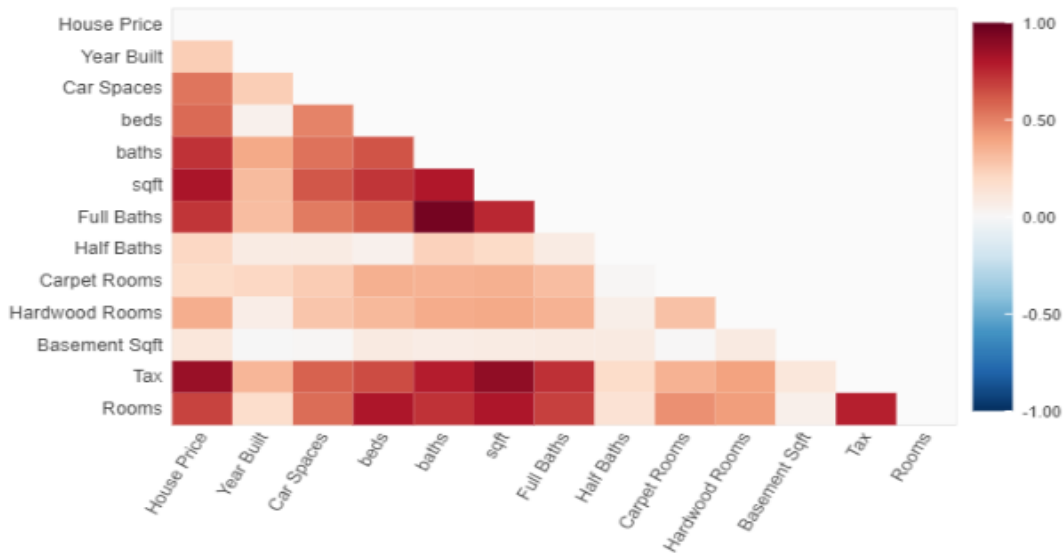

**Figure 1.** Heatmap of Housing Variables
Squares with a darker shade of red indicate a relationship between variables. The variables of tax, square footage, baths, and number of rooms are notable in their correlation with House Price.

## Modeling Methods

To model the data, five models were chosen, being linear regression, support vector regression, random forest regression, decision trees, and XGBoost regression. Hyper parameter tuning was also applied onto the models in order to minimize error. This modeling process was possible through the scikit Python package, which handled training and testing the models. The section below details important features for each model. Linear regression involves the relationship between a single dependent variable with other independent variables. Each input value is assigned a coefficient and an intercept to allow a higher degree of freedom. When there are multiple variables like there are in our analysis, linear algebra calculations are done using data matrices to estimate optimal values for input coefficients in order to minimize the sum of squared residuals.

*Linear Regression:*
Linear regression involves the relationship between a single dependent variable with other independent variables. Each input value is assigned a coefficient and an intercept to allow a higher degree of freedom. When there are multiple variables like there are in our analysis, linear algebra calculations are done using data matrices to estimate optimal values for input coefficients in order to minimize the sum of squared residuals.

*Support Vector Regression:*
Support vector regression builds upon linear regression by expressing tolerance towards error. The idea of SVR is to find the best fit line with hyperplanes, which are decision boundaries used to predict continuous output. The data points on each side of the hyperplane are known as Support Vectors. To define hyperplanes in higher dimensions, kernels are used, which are a set of mathematical functions that take data and transform it into the required state. Examples include linear, non-linear, and polynomial kernels. Finally, two lines drawn around the hyperplane create a margin of error. The goal of the model is to minimize this error.

*Decision Trees:*
This approach uses a tree structure with three types of nodes. The root node comprises the entire dataset and is further split during analysis. Interior nodes express the features of a dataset, and the results are shown by branches off the node. Finally, the leaf nodes show the numerical output. The algorithm involves making decisions and producing nodes in a fashion that reduces the error for the resulting nodes. The tree keeps splitting until the coefficient of variation reaches under a specified level, which is the standard deviation of the target variable divided by the mean of the target variable. This strategy is prone to overfitting, where the tree fits perfectly towards the sample dataset and thus has difficulty in predicting values outside of the sample data.

*Random Forest Regression:*
Random forest regression involves building off of the decision tree framework. Random forest uses a combination of two techniques, bootstrapping and ensemble learning, to improve the decision tree framework. Ensemble learning involves constructing multiple independent models, averaging the results to yield a more powerful result. Bootstrapping involves iterations of randomly sampling subsets of data, selecting a certain number of variables. Again, a more powerful result is calculated through the averages of these iterations.

*XGBoost Regression:*
XGBoost stands for extreme gradient boosting. This is another technique that improves decision trees. In this type of ensemble learning, sequential models are constructed out of subsets of the data where the output model serves as the input for the next construction, aiming to correct previous errors. This is done by weighting responses of models, where errors are weighted more heavily. The model then adjusts the weight of these errors on a gradient. XGBoost is an optimized version of this gradient adjustment to a tree.

## Results

The results of each model were calculated, and are evaluated based on mean absolute error, root mean squared error, and a score (coefficient of determination). Mean absolute error is calculated by averaging the differences between the predicted and real values for all data points. The equation below represents this computation, where $y_i$ is the true value and $x_i$ is the predicted value.

$$MAE = \sum i = 1^n |x_i - y_i|$$

The root mean squared error is measured by the square root of the standard deviation of the residuals, demonstrated by the formula below.

$$RMSE = \sqrt{\frac{\sum_{i=1}^{n}(y_i - x_i)^2}{N}}$$

Finally, the score, or coefficient of determination evaluates the proportion of variance in the dependent variable that can be explained by the independent variables. In other words, it represents how well the data fits the regression model, also known as goodness of fit. The calculation is detailed this equation, with $\overline{y}_i$ representing the mean value of the sample.

$$r^2 = 1 - \frac{\sum_{i=1}^{n}(y_i - x_i)^2}{\sum_{i=1}^{n}(y_i - \overline{y}_i)^2}$$

From the table, it can be seen that the XGBoost model overall performed the best, followed by random forest regression. XGBoost's results have generally been the most robust for every year, with the exception of 2018 where random forest regression was the most accurate by a small margin. In the past year (2021-2022), although error has remained proportionally similar, the value has increased. This is due to the increase in volatility and home price for the local real estate market. Details regarding the results of the models can be seen in table 2.

|  | Year | RMSE | MAE | R-square Score |
|---|---|---|---|---|
| **Linear Regression** | 2018 | 87655.28 | 56952.2 | 0.71 |
|  | 2019 | 98320.49 | 59334.61 | 0.71 |
|  | 2020 | 90084.82 | 62031.55 | 0.76 |
|  | 2021-22 | 106626.7 | 71518.17 | 0.72 |
| **Support Vector Regression** | 2018 | 88236.47 | 55099.18 | 0.71 |
|  | 2019 | 102863.2 | 61790.63 | 0.68 |
|  | 2020 | 91503.51 | 61708.86 | 0.75 |
|  | 2021-22 | 112782.6 | 71994.52 | 0.69 |
| **Random Forest Regression** | 2018 | 68536.04 | 42886.35 | 0.83 |
|  | 2019 | 78645.87 | 49254.93 | 0.81 |
|  | 2020 | 81160.84 | 51598.78 | 0.8 |
|  | 2021-22 | 87421.52 | 57972.93 | 0.81 |
| **Decision Tree Regression** | 2018 | 87912.62 | 56137.89 | 0.76 |
|  | 2019 | 90006.63 | 58319.38 | 0.71 |
|  | 2020 | 93989.22 | 60598.27 | 0.77 |
|  | 2021-22 | 103289.8 | 69797.02 | 0.83 |
| **XGBoost Regression** | 2018 | 70892.48 | 44079.3 | 0.81 |
|  | 2019 | 60609.87 | 41331.89 | 0.89 |
|  | 2020 | 77243.79 | 49889.3 | 0.82 |
|  | 2021-22 | 82072.88 | 53769 | 0.84 |

**Table 2.** Performance of Models in Dollars
The results of each model's performance can be seen. The columns are divided by the metric used to evaluate error. The results are also separated by year. XGBoost modeling yielded the best results.

## Variable Importance

Given the successful results of the XGBoost model, further analysis was conducted on the importance of variables in the results. Using SHAP, figure 2 details the importance of these variables on model output. SHAP is a technique used to evaluate variable importance by considering all possible combinations of variables when modeling. The results can then be compiled in graphs to display the weighted variables. As expected, square footage and annual tax amount have consistently assisted in calculation for property price as discussed earlier. Bathrooms and number of rooms tend to be strong as well, however their rankings tend to shift around a bit unlike square footage and tax which remains in the top two most important variables. A trend also exists where condo property type reduces appraisal price and single family residential homes increase price.

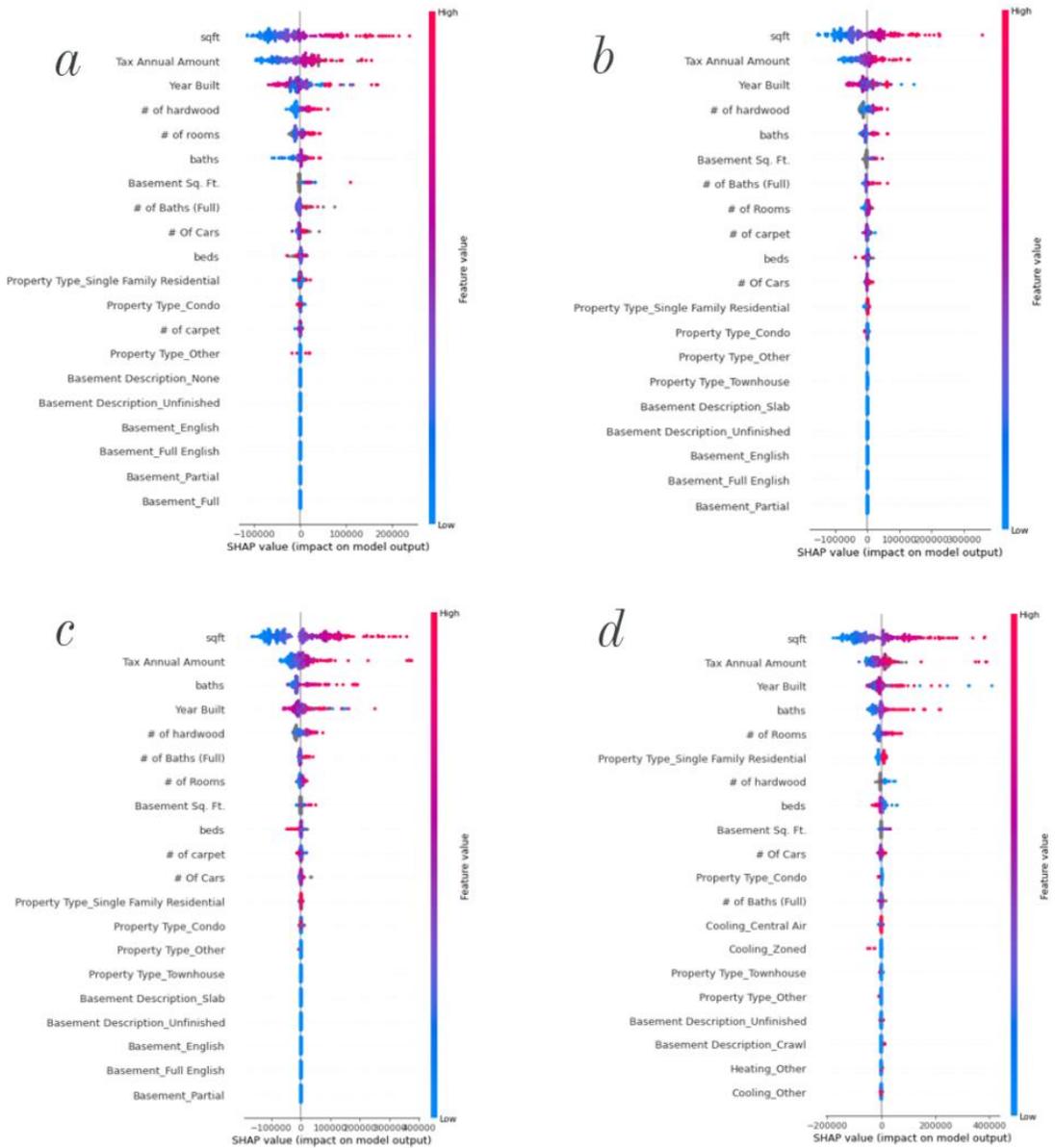

**Figure 2.** Shapley Values by Year
Figures 2a, 2b, 2c, and 2d show corresponding years of 2018, 2019, 2020, and 2021-2022 respectively. The SHAP values demonstrate that higher square footage and tax amount contribute greatly to model prediction. High and low values can be distinguished by the color gradient in the legend.

# Conclusion

In this study, five algorithms were used to estimate a property price based on attributes scraped from Redfin.com. The methods employed were linear regression, support vector regression, random forest regression, decision trees, and XGBoost regression. It was found that the performance of the XGBoost regression was the most effective in estimating the price of a property. After reviewing the most effective model, it was found through SHAP that square footage, tax annual amount, number of bathrooms, and number of rooms were the most important variables in determining a prediction.

With these results, XGBoost modeling has proved to be successful in estimating prices even in the current state of the volatile post-COVID-19 market. The error level remains relatively stable from 2018 to the summer of 2022. Overall, real estate market analysis can be enhanced through these models, and local realtors can have a rough estimate of housing prices through machine learning.

# Limitations

In the future, estimates could be improved in some ways. First, older years of Redfin data do not include details like basement style or flooring in rooms. Some estimates could be improved with these missing features. Also, seasonality can be introduced into the model in order to reduce variation in pricing, as houses tend to be cheaper in the winter and more expensive in the summer as families can move before the school year begins. With these additions, a more accurate model could be constructed.

# Acknowledgements

I would like to thank Hieu Ngyuen for his support in realizing this project. He offered tremendous help during the data cleaning and modeling. His constant assistance gave me inspiration in completing this project.